\documentclass[preprint,review,12pt]{elsarticle}

\usepackage{amssymb}
\usepackage{amsmath}
\usepackage{graphicx}
\usepackage[retainorgcmds]{IEEEtrantools}
\usepackage[colorlinks]{hyperref}
\hypersetup{
  citecolor = {blue}
}
\usepackage{siunitx}
\usepackage{pdflscape}
\usepackage{enumitem}
\usepackage[normalem]{ulem}
\useunder{\uline}{\ul}{}
\usepackage{setspace} % \doublespacing
\usepackage[utf8]{inputenc}
\usepackage[linesnumbered, ruled, vlined]{algorithm2e}
\usepackage{rotating}
\usepackage{nicefrac}
\usepackage{xspace}
\usepackage{float}
\usepackage{caption}
\usepackage{subcaption}
\usepackage{rotating, lscape, longtable, tabu, booktabs, siunitx, multirow}
\usepackage[capitalise, noabbrev]{cleveref}
\usepackage{verbatim}
\usepackage{mathtools}

\usepackage{tkz-graph}

\usetikzlibrary{arrows,arrows.meta,shapes,decorations.pathmorphing, decorations.markings, positioning}
\let\svtikzpicture\tikzpicture
\def\tikzpicture{\noindent\svtikzpicture}

\def\name{{SAFER}}

\newcommand{\uset}[1]{\ifmmode\left\{\,#1\,\right\}\else\{\,#1\,\}\fi}
\newcommand{\ulst}[1]{\ifmmode\left[\,#1\,\right]\else[\,#1\,]\fi}
\newcommand{\upar}[1]{\ifmmode\left(\,#1\,\right)\else(\,#1\,)\fi}
\newcommand{\uioc}[1]{\ifmmode\left(\,#1\,\right]\else(\,#1\,]\fi}
\newcommand{\uico}[1]{\ifmmode\left[\,#1\,\right)\else[\,#1\,)\fi}

\journal{Open-World AI}

\usepackage{multirow}
\usepackage{array}

\usepackage{tabularx,tabulary,ragged2e}
\newcolumntype{m}{>{\RaggedRight}X}
\usepackage{adjustbox}

\def\name{{SAFER}}
\newcolumntype{s}{>{\columncolor{lightgray}} m{7 em}}
\begin{document}

\begin{frontmatter}

\title{\name: Situation Aware Facial Emotion Recognition}

\author[unicamp]{Mijanur Palash\corref{cor1}}
\ead{mpalash@purdue.edu}
%\ead[url]{http://iagoac.github.io/}

\author[label2]{Bharat Bhargava}
\ead{bbshail@purdue.edu}

\address[unicamp]{Purdue University, USA}

\cortext[cor1]{Corresponding author}

\address[label2]{Purdue University, USA}

\begin{abstract}
In this paper, we present SAFER, a novel system for emotion recognition from facial expressions. It employs state-of-the-art deep learning techniques to extract various features from facial images and incorporates contextual information, such as background and location type, to enhance its performance. The system has been designed to operate in an open-world setting, meaning it can adapt to unseen and varied facial expressions, making it suitable for real-world applications. An extensive evaluation of SAFER against existing works in the field demonstrates improved performance, achieving an accuracy of 91.4\% on the CAER-S dataset. Additionally, the study investigates the effect of novelty such as face masks during the Covid-19 pandemic on facial emotion recognition and critically examines the limitations of mainstream facial expressions datasets. To address these limitations, a novel dataset for facial emotion recognition is proposed. The proposed dataset and the system are expected to be useful for various applications such as human-computer interaction, security, and surveillance.

\end{abstract}


\end{frontmatter}

\section{Introduction}\label{intro}
Human emotion recognition (ER) has gained significant research interest in recent years, particularly in the field of Artificial Intelligence (AI). This is due in part to the growing demand for online and remote learning systems as a result of the Covid-19 pandemic, where ER can play a crucial role in maintaining a positive and engaging learning environment by tracking the emotional status of students. Additionally, ER has a wide range of applications in domains such as human-computer interactions~\cite{cowie2001emotion}, law enforcement and surveillance~\cite{clavel2008fear}, interactive gaming, consumer behavior analysis, customer service~\cite{li2019acoustic}, and health care~\cite{ali2016novel}, among others. 

\begin{table}[!ht]
\centering
\caption{A list of facial expressions corresponding to different emotion categories~\cite{emotion_table}}
\begin{tabularx}{\textwidth}{|l|m|}
\hline
\textbf{Emotion Type}  & \textbf{ Corresponding Facial Expression}  \\ 
 \hline

   Anger & Eyebrows pulled down, upper eyelids pulled up, lower eyelids pulled up, margins of lips rolled in, lips may be tightened\\
   \hline
   Fear &Eyebrows pulled up and together, upper eyelids pulled up, mouth stretched\\
\hline
Disgust &Eyebrows pulled down, nose wrinkled, upper lip pulled up, lips loose \\
\hline
Happiness& Muscle around the eyes tightened, “crows feet” wrinkles around the eyes, cheeks raised, lip corners raised diagonally\\
\hline
Sadness&Inner corners of eyebrows raised, eyelids loose, lip corners pulled down\\
\hline
Surprised &  Entire eyebrow pulled up, eyelids pulled up, mouth hangs open, pupils dilated\\
\hline
Contempt &    Entire eyebrow pulled up, eyelids pulled up, mouth hangs open, pupils dilated\\
\hline

\end{tabularx}
\label{face_feat}
\end{table}

\begin{figure}[]
\centering
   \includegraphics[width=1\linewidth]{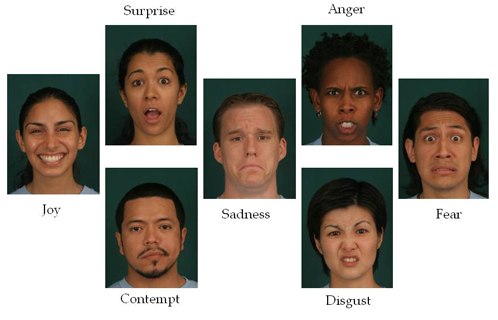}
   \caption{ Universal facial expressions of basic emotions~\cite{basic_emo}}
   \label{emo7}
\vspace{-0.1in}
\end{figure}

Facial emotion recognition (FER) is a widely adopted approach for ER, which primarily relies on the analysis of facial expressions to infer emotional states. Researchers have traditionally categorized basic emotions into seven distinct categories, including anger, happiness, sadness, disgust, fear, contempt, and surprise~\cite{patel2020facial}. Figure~\ref{emo7} illustrate some common facial expressions associated with each of these emotions. The universality of facial expressions across different cultures, as demonstrated in this well-known study at~\cite{elf}, has greatly facilitated the development of FER systems. Additionally, the presence of micro-expressions, which are involuntary facial actions indicative of concealed emotions, also play a critical role in FER.

The Facial Action Coding System (FACS) is a widely accepted method for describing the movements of various facial muscles associated with different emotions. FACS breaks down facial expressions into individual components of muscle movement, referred to as Action Units (AU). Table~\ref{face_feat} lists some common AU activities associated with different emotions. 

\begin{figure}[!t]
 \centering
\subfloat[]
{\includegraphics[width=0.36\linewidth]{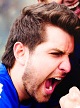}}
    % \caption{}
    %\label{sf1}
\hfill
\subfloat[]
{\includegraphics[width=0.545\linewidth]{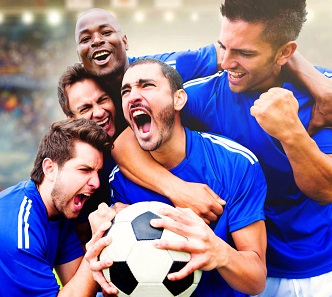}}
    % \caption{T}
   % \label{sf2}
\caption{Importance of background in emotion recognition. (a) Without background the subject appears angry. (b) The background context shows the subject is happy and celebrating a goal with his team-mates.}
\label{sf}
\end{figure}

Similarly, it is important to note that the situational context surrounding an individual also plays a significant role in shaping their emotional state. For instance, a person working in a sweaty coal mine is more likely to exhibit unhappiness as compared to someone walking in a park with their dog. However, in some situations, place type alone may not be sufficient. For example, in a stadium, depending on whether the team wins or loses, some people may be happy and some may be unhappy. Figure ~\ref{sf}(a) and \ref{sf}(b) illustrate this point, where without the background, the facial expression may be misleading. Therefore, the ability to extract and incorporate situational information, such as scene background and location type, can greatly enhance the accuracy of FER systems.

However, during the Covid-19 pandemic, the widespread use of face masks has presented a unique challenge for FER, as masks obscure facial expressions and result in a loss of important information. This can lead to a significant decrease in performance for models trained on datasets without masked subjects, with accuracy drops of up to 29\% observed in masked test sets. In this context, situational knowledge becomes critical, and special datasets and models that can deal with masked subjects are needed.

Moreover, the field of explainable AI has gained significant momentum in recent years~\cite{xai}, as the lack of transparency and interpretability of black-box deep learning systems has become a major concern. However, current ER works do not focus on this issue, and only classify the emotional status of the subject without providing any explanation for their decision. To address this, we propose the use of facial data, scene background, and place type, to create situational knowledge that can explain the results of our emotion recognition system, SAFER.

The main contributions of this work are:
\begin{enumerate}

    \item The development of a novel multi-stream emotion recognition system, SAFER, that utilizes deep learning methods to classify emotions from facial expressions, scene background, and location type.

    \item Evaluation of the proposed system on several benchmark datasets, and a comparison of our results with other recent works. Additionally, an ablation study is conducted to evaluate the effects of each stream in the learning process.
    
    \item A discussion of the drawbacks of existing FER datasets, including issues related to data quality, imbalanced class distribution, and racial and gender bias. 
    
    \item  A novel dataset for FER called DeFi, which includes masked subjects, to enable researchers to investigate the effect of masks on emotion recognition. 

\end{enumerate}

Overall, this work represents a significant step forward in the field of explainable artificial intelligence, by providing a transparent and reliable system for emotion recognition that utilizes multiple streams of data and context, and by addressing the challenges posed by the use of face masks. This work can be considered as a step towards open world AI, where AI systems can adapt to the changing scenario and can provide explanations for their predictions.

\begin{figure*}[htb]
\centering
   \includegraphics[width=1.0\linewidth]{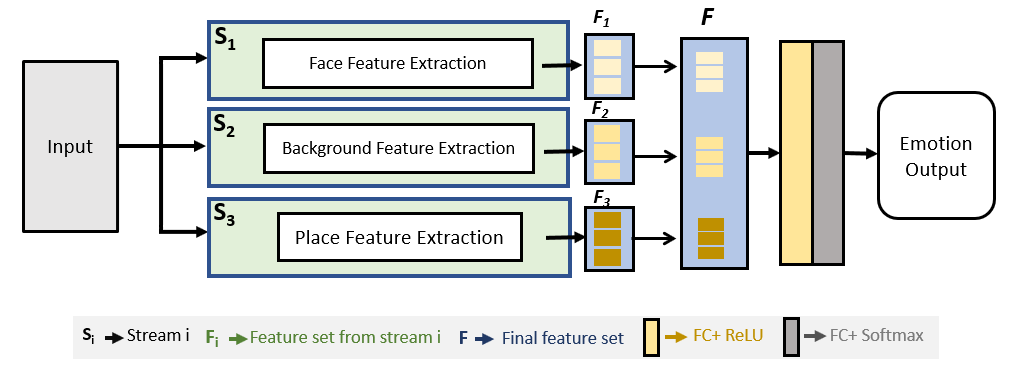}
   \caption{The \name\space system architecture consists of three parallel streams working on the input image for face, background and place feature extraction. A 2-layer deep neural network (DNN) performs emotion classification from the combined final feature set $F$.}
   \label{fig:sysD1}
\vspace{-0.1in}
\end{figure*}

\section{Background and Related Work}\label{background}
Emotions are a natural response to environmental stimuli. Understanding human emotions is crucial for effective communication. Human perception and judgment of situations or individuals are heavily influenced by their emotional state, which can impact activities such as driving a car, learning in a classroom, or interactions with law enforcement, among others.

%It can be defined as a class of feelings different from other sensory experiences such as tasting chocolate or preconceptions such as sensing pain in one’s lower back~\cite{stanford}. Feelings that are transient such as relief, longer-lasting such as mood, or long-lasting such as temperament all are part of emotion.

%The two most widespread theories in psychological research indicate that emotions are typically characterised in two broad categories: 1) emotions as discrete entities;\cite{ekman}) and 2) emotions as a combination of independent dimensions. 

% The former category is also known as basic emotion theory suggests that humans endowed with a limited set of emotions due to evolutionary reasons. Thus the behavioural and physiological expression of each emotion should be separate and arise from unique neural substrates~\cite{ekman}. Researchers of the latter category define emotions as a combination of two main dimensions: valence and arousal~\cite{carroll}.  
% \begin{itemize}
%     \item Valence: It is a continuum that varies from negative to positive values.  Valence is defined as the level of pleasantness that an event generates.
%     \item Arousal: Arousal is the level of autonomic activation that an event creates, and ranges from calm (or low) to excited (or high). 
% \end{itemize}
 
%  In this work, we take the first approach for emotion categorization and classify emotions into various distinct categories such as Anger, Fear, Disgust, Happiness, Sadness etc.    

In the field of facial expression recognition (FER), it is crucial to accurately detect and segment the facial region from the surrounding image. Various techniques have been proposed for this task, such as the Viola-Jones method~\cite{jayal,Li}, the combination of PCA and Viola-Jones~\cite{kar}, and the Haar Cascades method~\cite{shah}. These approaches have demonstrated effectiveness in accurately localizing the facial region for further processing in FER systems. A recent technique proposed by Bazarevsky et al.~\cite{blazeface} is capable of detecting six landmarks on a face and can handle the detection of multiple faces within an image.

%A convolutional neural network (CNN) is a type of deep learning architecture widely used in image analysis. The convolution is a repeated filter operation to an input that results in a map of activations called a feature map. It highlights the locations and strength of a detected feature in the input. CNN highlights important parts of an image which differentiate the classes. In~\cite{jadhav}, authors used a convolutional neural network (CNN) to detect emotion from the facial expressions on the FER-2013 dataset~\cite{FER2013}. 

Convolutional neural networks (CNNs) are a widely-utilized deep learning architecture in image analysis. The convolution operation, which applies repeated filters to an input, results in a map of activations known as a feature map. This highlights the locations and strength of features detected in the input, allowing CNNs to identify important parts of an image that differentiate between classes. In a previous study~\cite{jadhav}, the authors employed a CNN to detect emotions from facial expressions in the FER-2013 dataset~\cite{FER2013}.

Gan et al.~\cite{gan} achieved improved accuracy on the FER-2013 dataset using ensemble CNN and a novel label level perturbation strategy. In~\cite{dhankhar}, authors used the ensemble method and transfer learning with VGG16 and ResNet-50. Fard et al.~\cite{ac} proposed adaptive correlation-based loss for facial expression in the wild. Similarly, Farzaneh et al.~\cite{farzaneh} proposed an deep attentive center loss for facial expression recognition. Both works use variations of Deep Metric Learning (DML) for emotion recognition. Li et al.~\cite{gaCNN} proposed CNN with attention for occlusion aware facial expression recognition. Wang et al.~\cite{ran} used region attention networks for more pose and occlusion robust recognition. She et al.~\cite{dmue} used latent distribution mining and pairwise uncertainty estimation for facial emotion recognition.

Support Vector Machine (SVM) is another type of machine learning model. It tries to identify the largest margin plane between the classes. The SVM is also popular in facial emotion recognition due to its lightweight architecture compared to CNN. Authors at~\cite{datta} used SVM for emotion classification. Their method achieved 91.8\% test accuracy on the CK+ dataset. Deep Belief Network (DBN) is also used in emotion recognition. Authors in~\cite{kurup} reported 98\% accuracy on the CK+ dataset using the DBN technique. 

Bias in machine learning is an important challenge~\cite{bias1,bias2}. Buolamwini et. al.~\cite{bias2} showed that machine learning algorithms can discriminate based on classes like race and gender. Zeng et al.~\cite{face2exp} showed the bias in the class distribution of FER training datasets. He proposed a circuit feedback mechanism to tackle the issue.

 The Places Database~\cite{places} has a massive collection with 10 million labeled scene photographs from around the world. Additionally, it offers various pre-trained CNNs (Places-CNNs) for scene classification, which can be used for scene category and attribute identification.
 
 %In~\cite{anp}, the authors present a Visual Sentiment Ontology (VSO) database which has more than 3,000 Adjective Noun Pairs (ANPs). SentiBank, a visual concept detector library is also proposed. SentiBank can be used to detect the presence of various ANPs in any image.

\section{Datasets}\label{datasets}
In the field of facial emotion recognition, various datasets have been used to evaluate the performance of different algorithms. We utilized various datasets including FER-2013~\cite{FER2013}, AffectNet~\cite{affectnet} and RAF-DB~\cite{rafdb} for our experiments.  

The \textbf{AffectNet} is a comprehensive dataset of facial expressions that were sourced from the Internet via the use of 1,200 keywords related to emotions. This database comprises over one million facial images, with 440,000 of them being manually annotated. The images in this dataset are classified into eight distinct emotion categories and also include valence-arousal annotations.

The \textbf{FER2013} includes 28709 training images, 3589 validation images, and 3589 test images that are classified into seven different emotion classes. These images are posed and there is an observable class imbalance issue within the dataset, as the `Disgust' class only comprises 700 images, while some other classes have around 5000 images. This presents a challenge for accurate emotion recognition using this dataset.

The \textbf{CAER-S} dataset is a collection of nearly 70,000 facial expressions from 79 different TV shows. These data are manually annotated into six distinct emotion categories. 

The \textbf{RAF-DB} is a dataset containing 29,672 facial expression images. These images have been collected from the internet and manually labeled by 40 annotators, and are classified into 7 classes of basic emotions and 12 classes of compound emotions. The images in this database vary greatly in terms of subject age, gender, and ethnicity.

The \textbf{DeFi} is a new dataset proposed in this work. It contains 21,000 images that have been collected in both posed and wild settings. These images have been classified into seven basic emotion classes. Additionally, the dataset includes data that can be used to train an emotion recognition model to identify emotions even when the user is wearing a face mask. The specifics of this dataset will be discussed in Section ~\ref{newdataset}.

A summary of all these datasets is presented in table \ref{datasets}. 

\begin{table}[]
\centering
\caption{Summary of the various Facial Expression Recognition (FER) datasets, with N representing Neutral, S representing Sadness, Sr representing Surprise, H representing Happiness, F representing Fear, A representing Anger, B representing Boredom, P representing Puzzlement, Ax representing Anxiety, C representing Contempt, and D representing Disgust.}
\begin{tabularx}{\textwidth}{|l|l|l|m|m|l|}
\hline
 \textbf{Name}& \textbf{No of Items}   &\textbf{Type}& \textbf{Setting}  &\textbf{Classes}&\textbf{Author}  \\ 
 \hline
 CK+ & 593&Video&Posed and spontaneous&N, S, Sr, H, F, A and D& \cite{ck}\\
  \hline
   FER-2013& 35,887 &   Image& Posed& N, S, Sr, H, F, A and D& \cite{FER2013} \\
    \hline
  Emotic & 23,571  &   Image& Wild  &N, S, Sr, H, F, A, D and 19 other classes & \cite{kosti2019context} \\
   \hline
   AffectNet& 450,000  &    Image& Wild  & N, S, Sr, H, F, A, D and C &\cite{affectnet} \\
    \hline
   RAF-DB&29672&Image&Wild&N, S, Sr, H, F, A and D&\cite{rafdb}\\
    \hline
   CAER-S & 70,000     & Image& TV shows&N, S, Sr, H, F, A and D&\cite{caers}\\
    \hline
      FABO & 206     & Video& Posed&N, S, Sr, H, F, A, B, P, Ax and D&\cite{gunes2007bi}\\
       \hline
DeFi & 21,000     & Image& Posed and wild&N, S, Sr, H, F, A and D&This work\\

\hline
\end{tabularx}
\label{datasets}
\end{table}

\section{Our method: \name} In this section, we present our proposed emotion recognition system \name. Figure~\ref{fig:sysD1} shows a high-level diagram of the system and its components.  

\subsection{Input} The input of this system is an RGB image that contains the face and the background. The image can be a still photo, a frame from video footage, or a live video feed for continuous monitoring.

\subsection{Face Detection}\label{fd}
% \name \space needs to separate the facial area from other parts of the input image. For face detection, we leveraged the technique outlined in the works of Bazarevsky et el.~\cite{blazeface} called Blazeface. It can detect 6 landmarks on the face and supports the detection of several faces of an image.

The \name \space system requires the separation of the facial area from the rest of the input image. Face detection is accomplished through the use of the Blazeface technique, as described by Bazarevsky et al.~\cite{blazeface}. This method can detect six landmarks on the face and can handle the detection of multiple faces in an image.

 %An efficient implementation of this technique is discussed at~\cite{mp}. Blazeface uses a very compact feature extractor convolutional
%neural network designed specifically for lightweight object detection.% Instead of using $3\times 3$ convolution kernels everywhere along the model graph, they use $5\times 5 $ kernels. It reduces the computation time. However, in our experiments, we found it sometimes fails to properly crop the face area. For facial based recognition where we have to deal with low-quality and partial images frequently, this may be an issue we need to deal with. 

\begin{figure}[htb]
\centering
   \includegraphics[width=.9\linewidth]{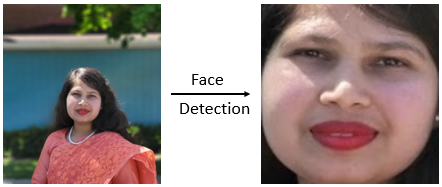}
   \caption{Facial area detection and separation}
   \label{fig:face}
\vspace{-0.1in}
\end{figure}

\begin{figure}[htb]
\centering
  \includegraphics[width=.9\linewidth]{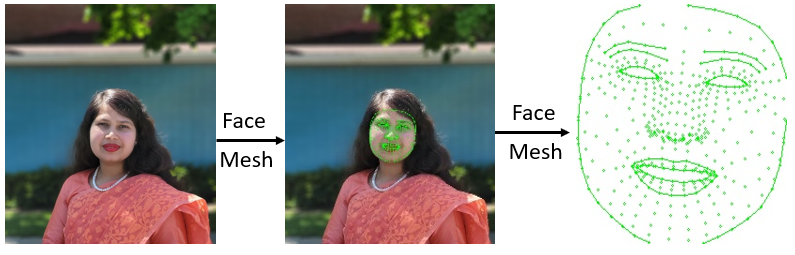}
  \caption{Face mesh generation.}
  \label{mesh}
\vspace{-0.1in}
\end{figure}

\begin{figure}[htb]
\centering
   \includegraphics[width=1.0\linewidth]{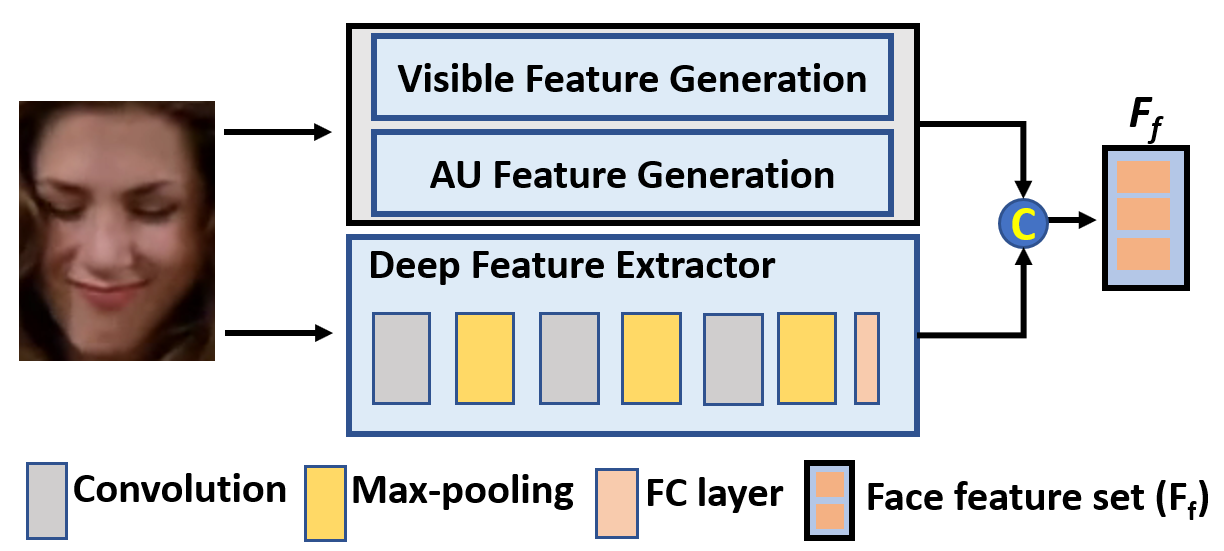}
   \caption{Face feature extraction. Three parallel modules work on face image to generate visible, action unit (AU) and deep feature sets. These are concatenated to create final face feature set $F_f$.}
   \label{face_f}
\vspace{-0.1in}
\end{figure}

\subsection{Face Feature Extraction}
This module consists of three components: the AU feature generator, the visible feature generator, and the deep feature extractor. These three feature sets are combined to produce the face feature set $F_f$. The different parts of this module are depicted in Figure~\ref{face_f}.

\subsubsection{Action Unit (AU) Feature Set}
From the face mesh (figure~\ref{mesh}) generated using Blazeface~\cite{blazeface}, we identify 12 key AU centers in the face
%We determine these centers based on the rules listed in table~\ref{au}. Then we choose the closest landmark positions as the centers. It simplifies the process without any noticeable change in accuracy. We determine the distances between all AU points which makes up our AU feature set. 
based on a set of predefined rules outlined in Table~\ref{au}. The centers are determined by selecting the closest landmark positions, which simplifies the process without sacrificing accuracy. Finally, we calculate the distances between all AU points to form our AU feature set. 

\begin{table}[]
\centering
\caption{Rules to select 12 action unit (AU) centers of the face.}
\begin{tabular}{|l|l|l|}
\hline
AU ID& Name & Rule  \\ 
 \hline
 1& Inner Brow Raiser & Above inner brow \\
 2&Outer Brow Raiser & Above outer brow\\
 4&Brow Lowerer& At brow center\\
 6& Cheek Raiser & At cheek center\\
 7& Lid Tightener & Top eye lid center\\
 10& Upper Lip Raiser & Upper lip center\\
 12 & Lip Corner Puller & Lip corner\\
 14& Dimpler& Below lip corner\\
 15& Lip Corner Depressor& Lip corner\\
 17 & Chin Raiser & chin center\\
 23& Lip Tightener& Bottom lip center\\
 24& Lip Pressor & Bottom lip center\\
 \hline
 \end{tabular}
 \label{au}
 \end {table}

\subsubsection{Visible Feature Set}
From the face mesh generated using Blazeface~\cite{blazeface}, we create a group of features shown in the Table~\ref{vis_feat}. 
They capture various aspects of the face including width, height, distance, and angle of different facial parts. Our approach leverages the observation that changes in facial expression, such as when shouting or laughing, often result in alterations to specific facial features. For example, when smiling or laughing, the mouth tends to open, leading to increased lip width. Conversely, expressions of surprise often result in increased eye width and height. These features are thus valuable for use in facial emotion recognition.

%These features deal with the width, height, distance and angle of different face parts. While shouting or laughing, people open their mouths resulting in a larger width (separation between the lips). Similarly, a surprised person will have big open eyes resulting in bigger eye width and height. Therefore, these features are valuable for our facial emotion recognition purpose. 

\begin{table}[]
\centering
\caption{Description of visible features of the face.}
\begin{tabular}{|l|l|}
\hline
Feature Type  & Feature Description  \\ 
 \hline

  \multirow{3}{*}{Width}& Left Eye\\
   
   &Right Eye\\

& Mouth\\
\hline
\multirow{4}{*}{Height}&Right Eye\\

&Left eye\\

& Right eye\\

&Mouth\\
\hline
 \multirow{5}{*}{Distance} & Left and right eyes\\
   \
   &Eyes to brows\\

& Eyes to mouth\\

& Eyes and nose\\

&Nose and mouth\\
\hline
\multirow{4}{*} {Angle} &Left eye with right eye and mouth\\

&Right Eye with left eye and mouth\\

&Mouth with both eyes\\

&Mouth with both eyes\\
\hline

\end{tabular}
\label{vis_feat}
\end{table}

\subsubsection{Deep Feature Set}~\label{deep_f}
Deep features are the values we obtain from the output of the deep feature extractor as shown in figure~\ref{face_f}.  For the feature extractor we experiment with two different CNN model types. 
\begin{itemize}
    \item Regular CNN: In the regular CNN setup, we use three convolutional layers and one fully connected layer. Input images of different resolutions are converted to $226\times226$ . All convolutions have a filter size of $2\times2$ and a stride of 1. %We use a stride of 1 in convolution because smaller strides work better in our experiments. It allows leaving all spatial downsamplings to the downsampling layers. All downsampling layers use $2\times2$ max-pooling technique. 
    The use of a stride of 1 in convolution was found to be more effective in our experiments as it permits all spatial downsamplings to occur in the downsampling layers. The downsampling layers employ the $2 \times 2$ Max-Pooling technique.
    
    \item Transfer learning from ResNet-50: 
    In this configuration, the Transfer Learning approach is utilized for deep feature extraction. This approach addresses the challenge of an insufficient number of training data points by transferring knowledge from a model that has been trained on a large dataset with similar properties to the smaller dataset in question. In this study, the ResNet-50 model~\cite{he2015deep} trained on the ImageNet~\cite{imagenet} database is employed. ResNet-50 is a deep neural network with 50 layers, which can mitigate the vanishing and exploding gradient problems that are commonly encountered in substantially deeper neural networks through the utilization of a deep residual learning framework and residual mapping technique.

\end{itemize}

\begin{figure}[htb]
\centering
   \includegraphics[width=1.0\linewidth]{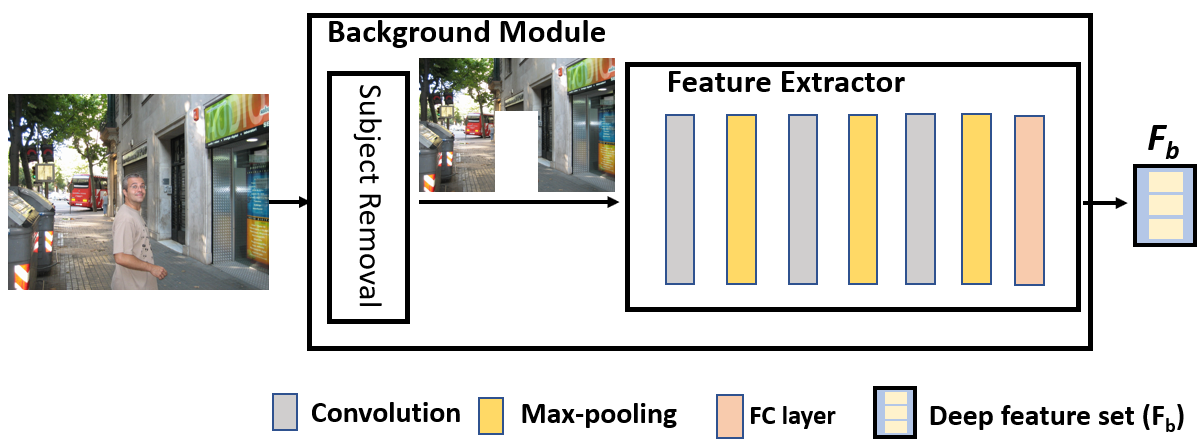}
   \caption{Background feature extraction module. Subject face and body is removed from the image and passed through the CNN network in this step. }
   \label{bg_1}
\vspace{-0.1in}
\end{figure}

\subsection{Background Feature Extraction}
We first remove the subject body and face from the scene to extract background. The extracted background is then processed through a deep feature extraction network as depicted in Figure~\ref{bg_1}. The network consists of three convolutional layers and one fully connected layer, with all convolutional layers utilizing a filter size of $2 \times 2$ and a stride of 1. The downsampling layers employ the $2 \times 2$ Max-Pooling technique. The result of this network is the Background Feature Set ($F_b$).

%Then it is passed through a deep feature extraction network shown in figure~\ref{bg_1}. We use three convolutional layers and one fully connected layer. All convolution layers use a filter size of $2\times2$ and a stride of 1. All the downsampling layers use $2\times2$ max-pooling technique. We get the background feature set ($F_b$) as the output of this network. 

\subsection{Place Feature Set}
In this step, we remove the subject body and face from the scene. We use pre-trained AlexNet provided with the Places dataset and pass the scene through it. Deep features are collected from AlexNet after the final max-pooling operation to produce the Location Feature Set ($F_l$). We also collect final place categories such as `classroom' and attributes such as `no\_horizon' and `enclosed\_area' for explanation generation. 

\subsection{Detection Model}
 The final feature set ($F$) is a concatenation of the face feature set ($F_p$), background feature set ($F_b$) and place feature set ($F_l$). We use two FC layers (figure~\ref{fig:sysD1}) for the final classification. Cross-entropy loss is used for the loss function.

\section{Experimental Results}\label{result}
%In this section, we discuss the experimental evaluation of \name \space in different facial emotion recognition datasets. We train our model with the benchmark datasets listed in table~\ref{datasets}. We compare our performance with some of the recent approaches proposed in the literature. We perform the experiments on a server PC. This PC has 20 core 2.6 GHz Intel Xeon CPU and 96 GB of memory. It also has 3 NVIDIA TESLA GPUs with 24 GB memory each. For accelerated computing, we use python multiprocessing and mixed precision libraries. 
%We split the datasets using an 80:10:10 ratio into training, validation and test sets. All images are re-scaled to $224\times 224$ pixels. We perform dataset augmentation using crop, rotation, brightness and contrast. We use an adaptive learning rate starting at $1e^{-5}$ and a batch size of 32. 

In this section, we present the experimental evaluation of \name \space on various facial emotion recognition datasets. Our model is trained on the benchmark datasets listed in Table~\ref{datasets}, and its performance is compared to recent approaches in the literature. The experiments were conducted on a server PC that had 20 cores with a 2.6 GHz Intel Xeon CPU and 96 GB of memory, as well as three NVIDIA TESLA GPUs with 24 GB of memory each. To facilitate accelerated computing, we used Python multiprocessing and mixed precision libraries.
The datasets were split into training, validation, and test sets in an 80:10:10 ratio. All images were resized to $224 \times 224$ pixels, and dataset augmentation was performed using cropping, rotation, brightness, and contrast adjustments. We employed an adaptive learning rate that started at $1e^{-5}$ and a batch size of 32

We use test accuracy as our performance criteria. The test accuracy is given by the following equation:

\[Accuracy=\frac{\#N_c}{\#N_t}\]

Where $\#N_c$ indicates the number of items correctly predicted and $\#N_t$ indicates the total number of items in the test dataset. 

\begin{table}[]
\centering
\caption{\name \space performance on various emotion recognition datasets}

\begin{tabular}{|l|l|}
\hline
\centering
 \textbf{Dataset} & \textbf{Test Accuracy(\%)}    \\ 
 \hline
  CK+& 98.5 \\
   FER-2013& 75.8 \\
   AffectNet&62.1 \\
   RAF-DB&87.1\\
   CAER-S & 91.4\\
   FABO & 96.1    \\

\hline
\end{tabular}
\label{result_datasets}
\end{table}

\begin{table}[]
\centering
\caption{\name \space performance comparison on CAER-S dataset.}

\begin{tabular}{|c|c|c|}
\hline
\centering
 \textbf{Method} &\textbf{Year}&\textbf{Test Accuracy(\%)}    \\ 
 \hline
  Lee et al.~\cite{lee2019context}&2019& 73.51 \\
   Kosti et al.~\cite{kosti2019context}&2019& 74.48 \\
   Li et al.~\cite{li2021human} &2021& 84.82 \\
   \name&2022&91.4\\

\hline
\end{tabular}
\label{result_caers}
\end{table}

\begin{table}[]
\centering
\caption{Performance comparison of different emotion recognition systems on FER-2013 dataset }
\begin{tabular}{|l|l|l|}
\hline
 \textbf{Method}&\textbf{Year}  &\textbf{Test Accuracy(\%)}  \\ 
 \hline
   %Mollah et al.\cite{mollahosseini}& 2015 & 66.0  \\
  ECNN\cite{wen}& 2017  &   66.98   \\
   Dhankhar\cite{dhankhar} & 2019& 67.2 \\
   Renda\cite{renda}& 2019 &   71\\
   Gan\cite{gan}&2019& 73.73\\
   A-C\cite{ac}&2022&72.03\\
    %  \name&2022& CNN &  68.1\\
    %   \name&2022& Resnet-50 &  72.3\\
   \name&2022&   75.8\\
\hline
\end{tabular}
\label{t_acc1}
\end{table}

% \begin{figure}[!tb]
% \centering
%   \includegraphics[width=0.9\linewidth]{figs/heatmap_fer.PNG}
%   \caption{Confusion matrix of \name \space in FER-2013 dataset.}
%   \label{conf_mat}
% \vspace{-0.1in}
% \end{figure}

% \begin{figure}[!tb]
% \centering
%   \includegraphics[width=0.9\linewidth]{figs/heatmap_caer.PNG}
%   \caption{Confusion matrix of \name \space in FER-2013 dataset.}
%   \label{conf_mat}
% \vspace{-0.1in}
% \end{figure}

\subsection{Performance of \name \space on Benchmark Datasets}
The results of \name \space with some facial emotion recognition benchmark datasets are shown in the table~\ref{result_datasets}. Our results are comparable with the state-of-the-art works in all datasets. For the FABO dataset, we outperform the accuracy reported by various recent works (table~\ref{t_acc1}). For the CK+ dataset, we find the best-reported result to be 98.57\% accuracy as in~\cite{kurup}, our accuracy of 98.5\% is comparable to it.

Besides, we report the results of several recent works on the CAER-S dataset in table~\ref{result_caers}. We see \name\space outperforms closest result reported by Li. et al.~\cite{li2021human} by 7.8\%. 

In FER-2013 dataset (table~\ref{t_acc1}), our model outperforms results from ~\cite{ac} and ~\cite{gaCNN}. 
Similarly, table~\ref{result_afn} shows results from several recent works on the AffectNet dataset. Here, our accuracy is 63.7\%. Which is close to the accuracy values reported by~\cite{face2exp} and ~\cite{farzaneh}. In~\cite{face2exp}, author provided a effective solution to the class imbalance issue widespread in most of the FER datasets including AffectNet. In AffectNet dataset `Happiness' class has 146,198 samples while `Disgust' has only 5,264 samples. For this reason our model gives lower accuracy in `Disgust' class. However, our work is orthogonal to ~\cite{face2exp} and both can be implemented together. Similarly, \cite{farzaneh} uses Deep Metric Learning (DML) method with modified loss functions. They argued that using softmax loss can not provide proper discrimination between classes due to inter-class similarity and intra-class variations. Hence they use sparse center loss in adition to the softmax as the final objective function. This work is also orthogonal to our work as this loss function can be used with our method too. 

From table~\ref{result_caers} and ~\ref{result_afn}, it is clear that our model offers greater improvement in the CAER-S dataset than AffectNet. This is due to the presence of less class imbalance and more contextual information in the CAER-S dataset.

\begin{table}[]
\centering
\caption{\name \space performance comparison on AffectNet dataset.}

\begin{tabular}{|l|l|l|}
\hline
\centering
 \textbf{Method} &\textbf{Year}&\textbf{Test Accuracy(\%)}    \\ 
 \hline
   DMUE~\cite{dmue}&2021&63.11\\
  SCN~\cite{scn}&2020&60.23\\
   SL~\cite{he2015deep}&2016 & 58.27\\
   F2E~\cite{face2exp} &2022& 64.23    \\
   A-C~\cite{ac}&2022&63.36\\
 DACL~\cite{farzaneh}&2021&65.2\\
   \name&2022&63.7\\

\hline
\end{tabular}
\label{result_afn}
\end{table}

% \begin{table}[]
% \centering
% \caption{\name \space Performance comparison on RAF-DB}

% \begin{tabular}{|c|c|c|}
% \hline
% \centering
%  Method & Year& Test Accuracy(\%)    \\ 
%  \hline
%   gaCNN~\cite{gaCNN}&2018& 85.07 \\
%   DMUE~\cite{dmue}&2021& 88.76 \\
%   SCN~\cite{scn}&2020&87.03\\
%   SL~\cite{he2015deep}&2016 & 84.16\\
%   F2E~\cite{face2exp} &2022& 88.54 \\
%   A-C\cite{ac}&2022&86.96\\
%   Ours&2022&87.1\\

% \hline
% \end{tabular}
% \label{result_afn}
% \end{table}

\begin{figure}[!t]
 \centering
\subfloat[]
{\includegraphics[width=0.6\textwidth]{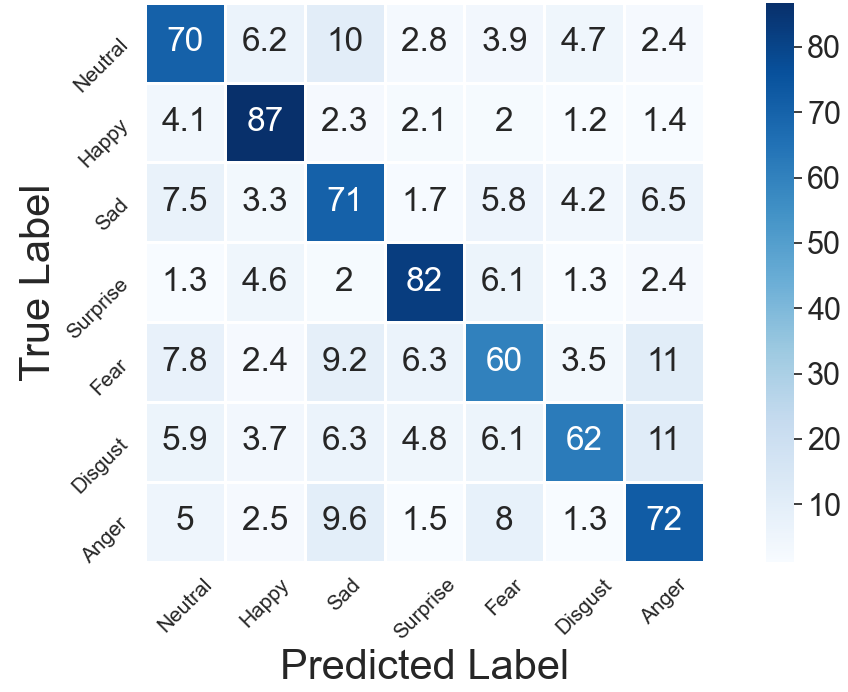}}

\subfloat[]
{\includegraphics[width=0.6\textwidth]{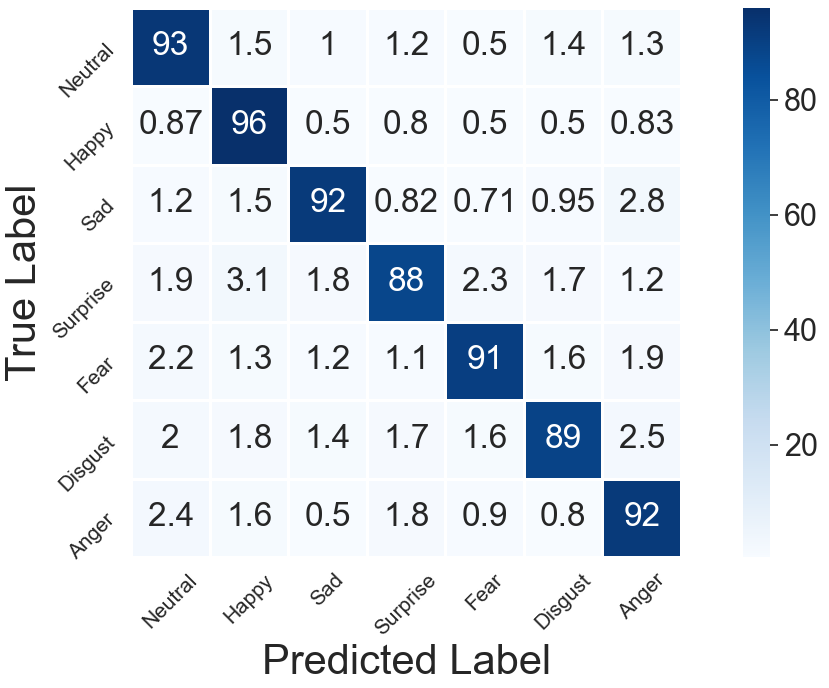}}

\caption{Confusion matrices of our model. (a) FER-2013 dataset (b) CAER-S dataset. Some of the classes in FER-2013 have a small number of samples, leading to significant accuracy differences among the classes. That is not the case for CAER-S and all classes have similar accuracies.  }
\label{conf_mat}
\end{figure}

Figure~\ref{conf_mat} shows the confusion matrices of our model for the FER-2013 and CAER-S datasets. In the figure, we list actual emotion labels along the vertical direction and predicted ones along the horizontal direction. An entry `$C_{ij}$' in row `$i$' and column `$j$' represents the number of samples who has the true label of row `$i$', and the predicted label of column `$j$'. 

From the confusion matrix of FER-2013, we can see some classes such as `Happiness' and `Surprise' are better recognizable while `Disgust' and `Fear' classes are not. But we do not see a similar pattern in the CAER-S dataset. One possible reason is the unequal distribution of samples in FER-2013 classes. For example, in FER-2013 the `Disgust' class has only 436 training samples while the "Happiness" class has 7215 training samples. But in CAER-S all the emotion classes have an equal number of samples (7001 samples). Hence, our model results in similar accuracy in all the emotion classes of the CAER-S dataset. 

Many of the emotion classes share some of the facial expression with each other. For example, we see lots of `Happy' samples are wrongly classified as `Neutral', `Disgust' as `Anger' etc. This happens due to the close correlation between these emotion classes and increases the complexity of the classification task. 

\subsection{Explanation Generation}

\begin{figure}[]
\centering
   \includegraphics[width=.9\textwidth]{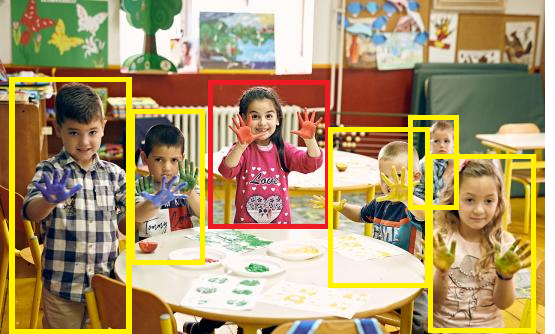}
   \caption{Explanation generation}
   \label{kinder}
\vspace{-0.1in}
\end{figure}

Besides determining emotion we also provide a guideline to generate an explanation for that result. We can provide human explainable reasoning by creating an idea of the situation around the subject. Individual modules tell us what information is available from the face and background. For instance, the subject in figure~\ref{kinder} red bounding box has a smiling face and colorful vibrant background. By extracting age, gender, location type and location attributes, we can create our situational knowledge which further enhances this reasoning. In the case of figure~\ref{kinder}, place category output is a day care play room. By combining all these a human understandable explanation of happiness class for the subject can be constructed as "the subject is a child in a playroom and smiling, has a happy facial expression". 

This is an early effort for explainable emotion classification using multiple data modalities as per our knowledge. However, our explanation generation still requires further work. Therefore, we are planning to explore explanation generation in a more detailed manner in our future works.

\subsection{Ablation Study}
We show the resulting outputs from various ablation experiments in table~\ref{ablation}. We compare the accuracy of our model on AffectNet and CAER-S datasets in several combinations. The combinations are: face feature set $F_p$ only, face feature set $F_p$ + background feature set $F_b$, face feature set $F_p$ + place feature set $F_l$ and all three feature sets combined. 

From the table, we can see that in the AffectNet dataset, operating on face data alone results in an accuracy of 61.9\%. Adding background and place feature sets does not improve accuracy significantly. But for the CAER-S dataset, we see the addition of these two extra feature sets result in a noticeable increase in accuracy. Our understanding is that the AffectNet dataset has only face images with a paltry background. They offer very little information for the background and place streams to extract. However, CAER-S is designed for context-based emotion recognition. Hence, we have lots of background information in the samples. That is why the contribution of background and place streams are significant for CAER-S.

We also analyse the choice of the deep feature extractor. In this case, pre-trained ResNet-50 performs better in our experiments than the regular CNN extractor discussed in section ~\ref{deep_f}.  We also notice that adding an AU feature set and visible feature has a positive effect on the accuracy of the face stream alone. If we remove them and use only the deep features from the face using ResNet-50, the accuracy drops to a lower value. 

Thus when contextual information is available, adding background or place features improves accuracy and the best accuracy result is achieved when we use all available feature sets. However, if no contextual information is available then these two modules fail to offer meaningful contributions and our model does not provide the best results.

\begin{table}[]
\centering
\begin{tabulary}{1.0\columnwidth}{|l|l|l|l|l|}
\hline
\textbf{F}&\textbf{B}&\textbf{P} &\textbf{AffectNet}&\textbf{CAER-S}\\
\hline
\checkmark&&&61.9&86.7\\

\hline
\checkmark&\checkmark&&62.01&89.3\\

\hline
\checkmark&&\checkmark&61.95&90.1\\
\hline
\checkmark&\checkmark&\checkmark&62.1&91.4\\
\hline
\end{tabulary}
\caption{Ablation study. Here, F, B and P stand for face, background and place feature set respectively. CAER-S images offer significant contextual information and B and P streams offer noticeable contributions in accuracy. However, AffectNet images lack contextual information leading to less contribution from B and P streams.}
\label{ablation}
\end{table}

\section{Issues of Facial Expression Based Emotion Recognition}\label{issues}
\subsection{Covid  Mask Issue}

During the global Covid-19 pandemic, people used to wear a face mask~\cite{cdc}. Face mask covers a good part of our face including the nose, lips, mouth and chin. This unexpected situation is a novelty and creates a challenge for facial-based emotion recognition, as the absence of these facial features can result in the loss of important cues for recognition. To examine the impact of masks on facial emotion recognition, a masked section has been included in our proposed DeFi dataset.

Testing on this mask dataset with a regular dataset-trained model gave us only 38.58\% accuracy. This highlights the need for a specialized dataset for masked subjects as models trained on regular face images are not effective at classifying emotions from partially visible faces.

We further trained and evaluated our model using the masked section of our proposed DeFi dataset. The results showed an increased accuracy of approximately 58\%, representing a 29\% improvement compared to the results from testing the regular dataset-trained model. Despite this improvement, the accuracy still remains lower compared to the results obtained from training and testing on unmasked data.

\begin{figure*}[htb]
\centering
   \includegraphics[width=.99\linewidth]{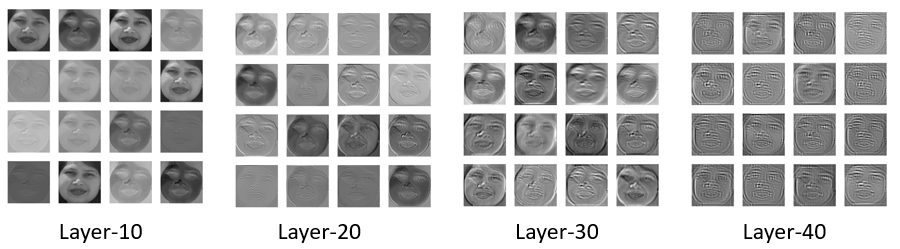}
   \caption{\name \space feature maps for an input image on different convolution layers highlighting which parts of the input image are getting higher importance on those layers of the model. It can be seen that the eyes, nose and lips are getting highlighted in most of the images. }
   \label{featuremap}
\vspace{-0.1in}
\end{figure*}

We further analyze this result with the feature maps from different convolution layers of the ResNet-50 of our architecture. The selected feature maps from layer 10, 20, 30 and 40 are shown in Figure~\ref{featuremap}. A visual inspection of these maps reveals that the lips and nose regions are highlighted in most of them. This highlights the fact that these facial parts are crucial for emotion classification, as feature maps highlight the most important parts of the image. As such, covering these important facial features through the use of masks significantly hinders the ability of facial expression-based models.

%These feature maps shown in figure~\ref{featuremap} are chosen from layer 10, layer 20, layer 30 and layer 40. From a visual inspection of these feature maps, we can see the lips and the nose area are highlighted in most of them. We know the feature map highlights more important parts of the image. Hence, lips, mouth and nose are important for our emotion classification.  Therefore, by covering those important facial parts, a mask makes facial expression-based models less capable.

% A possible solution to this challenge can be multi-modal emotion detection, where multiple modalities such as the face, posture, gait etc. are used for emotion recognition. However, these systems tend to be more complicated than only facial expression-based systems and they require different types of datasets such as posture and gait information. These are not available in most of the FER datasets. Multi-modal emotion recognition is beyond the scope of this work.

One potential approach to address the issue of face masks affecting facial expression-based emotion recognition is to employ multi-modal emotion detection, which leverages multiple modalities, such as facial expression, posture, and gait, to recognize emotions. However, these multi-modal systems tend to be more complex compared to single-modality facial expression-based systems, and also require diverse types of datasets that are not typically available in most FER datasets, such as posture and gait information. Thus, multi-modal emotion recognition is left as a future work and not addressed in this current study.

\subsection{Dataset Bias}
 A good learnt model is dependent on a good dataset. However, without proper care a dataset can lack proper diversity. The model trained on it then performs worse when it encounters minority subjects of the dataset. For example, keyword searching in Google with "angry face" resulted in 59 acceptable images in the first 100 images. Of these images, only 9 are women while 50 are men. This pattern holds for other generic keywords such as "sad people" or "happy human".  Preparing a dataset by collecting results from generic keywords instead of more specific ones such as "black woman sad" has a higher chance of bias against women and people of color.  A similar situation is also applicable to the volunteer choice for creating an acted dataset. Without the careful selection of people from various genders and ethnic backgrounds, dataset bias can be easily incorporated into the model. Table~\ref{bias1} shows the percentage of images with male subjects for different gender-neutral emotion-related keywords for the first 100 results listed by Google. As we can see, the number of male and female images is not equally represented in the results. For anger-related searches, men have a higher percentage of images while in sadness-related searches women tend to appear in larger numbers. We also show the number of males in the first 100 images in the FER-2013 dataset for the "Angry", "Fear", "Happy" and "Sad" categories and report similar and unequal male/female representations. 
 
 Besides gender and race, many of the existing FER datasets are biased toward majority classes. Some of the classes have a larger number of samples and hence models tend to show favor towards them. For example, in the AffectNet dataset, the `Happy' class has more than 130k samples and the `Disgust' and `Contempt' classes have only a few thousand samples each.

\begin{table}[]
\caption{Gender bias- number of images with male-looking subjects from gender neutral-keyword searches on Google, and on the FER-2013 dataset [first 100 images]. }

\begin{tabular}{|l|l|l|}
\hline

 \textbf{Keyword} &\textbf{\# Male in Google(\%)}&\textbf{\# Male in FER-2013}    \\ 
 \hline
  "Angry people"& 84.7 & 70\\
  "Fear face"& 60.1&52\\
   "Happy human face"& 55.8 &58\\
    "Sad human face"& 40.0 &45\\

\hline
\end{tabular}
\label{bias1}
\end{table}

\subsection{Quality Concern of Current Datasets}
In table~\ref{datasets} we listed some widely used datasets for facial emotion recognition. By having a closer look at them, we found certain issues we can work on resolving. Some of the images from CAER-S, Emotic and FER-2013 are shown in figure~\ref{fer_fault}. The main issues we found in these datasets are:
\begin{itemize}
 \item Noise issue: Some of the images are not relevant and contain no face. We suspect this is due to the Google image scrapping.  In our search, we find that Google includes some irrelevant images in the search results. We need to clean them manually. Examples of this type of image are shown in the first row of figure~\ref{fer_fault}.
 \item Quality issue: Some of the images are cropped and lack parts of the face/ full face as shown in the second row of figure~\ref{fer_fault}. As we can see from figure~\ref{featuremap}, we need all those parts of the face for successful recognition. 
 \item Confusing annotation: These are the images where annotations do not match with the image. This is certainly an issue for those images where there is a lack of consensus among human annotators. In some of the datasets where they used frames from video clips of movies and dramas, they annotated all the frames for a clip as a certain class, even though emotion changes from neutral to apex to back to neutral again.  Using a larger number of annotators per image and removing confusing images can resolve this issue. All the images on the third row of ~\ref{fer_fault} are annotated at angry. The first three are from a popular TV series and those particular situations can be best termed neutral conversations. 
 \item Number of items in classes: Another issue we found is having a large difference in the number of data points per class. For example, in FER-2013 the `Angry' class has 3995 images while the `Disgust' class has only 436 images. A good dataset should have an equal number of samples in each of the classes. 
\end{itemize}

\section{Proposed DeFi Dataset}\label{newdataset}
Recent works have proposed several algorithmic methods to deal with the above-mentioned bias in the datasets~\cite{face2exp,ac,bs1}. However, these methods do not mitigate the problem completely. Hence in this work, we intend to propose a new dataset where samples are carefully chosen to be more inclusive. This is an orthogonal approach to the algorithmic solutions and they can work together. 

\begin{figure}[!tb]
\centering
   \includegraphics[width=.95\linewidth]{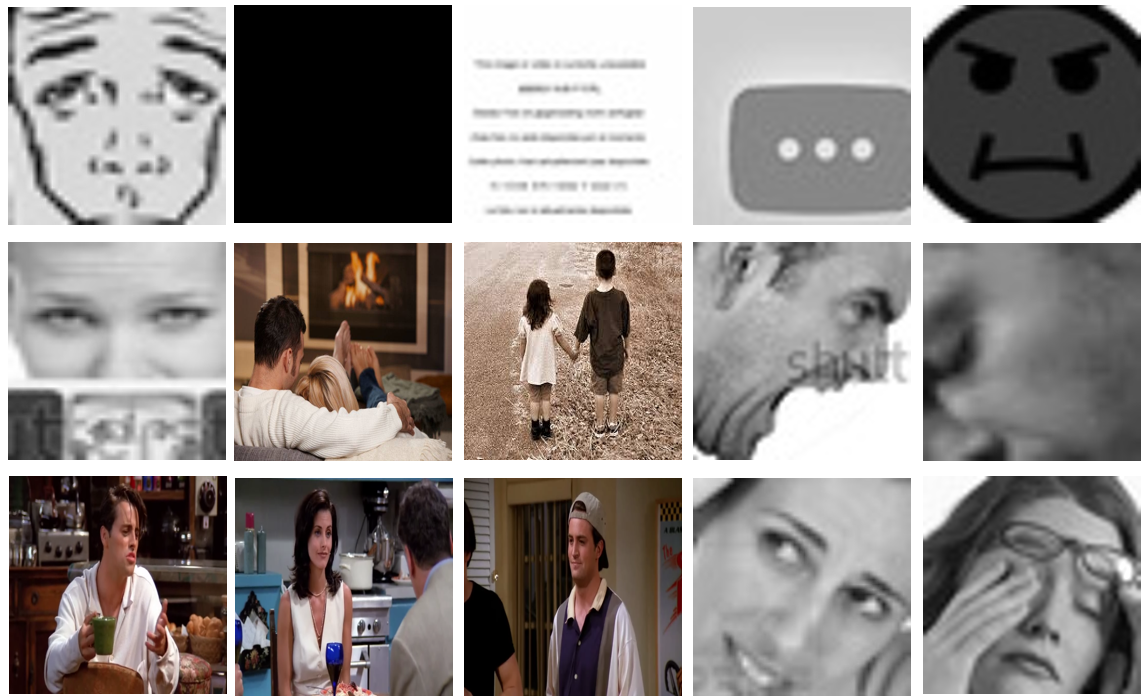}
   \caption{ Quality issues in current datasets. From top to bottom non-relevant images, images with missing facial parts and images with wrong annotations.}
   \label{fer_fault}
\vspace{-0.1in}
\end{figure}

% \subsection{Dataset }
We first start with improving the existing datasets according to the issues outlined above. We have 8 volunteers, 4 males and 4 females, all aged 18-30 years and university students, checking the datasets for any irrelevant images similar to the first row in the figure~\ref{fer_fault}. They also annotate the images individually. From these annotations, we keep images with at least 4 annotators agreeing on the label. We call it the 80\% consensus method. After these steps, approximately 7.5\% images are removed from the original FER-2013 dataset. From FABO dataset videos, we extract the frames at a 10 FPS rate. As the author outlined, subjects in this dataset go from neutral to apex emotion state and then back to neutral. Our volunteers collect the images showing apex emotion states from the extracted frames. We again follow 80\% consensus among the annotators for selecting an image in this step.

Besides working on existing datasets we record video clips showing 7 emotional states from a group of 10 volunteers. Due to the Covid 19 pandemic, we do not meet the participants in person, rather they are recruited over social media platforms. These volunteers are aged 18-30 and from both genders with diverse racial and ethnic backgrounds. All are university students. Each participant records video clips of their best impression of these emotions and provides us with the clips. We extract the video frames as 10 FPS and perform the 80\% consensus check by the annotators. 

To increase diversity in the dataset and reduce bias, we include people from multiple geographic and ethnic backgrounds by collecting images from Google searches using keywords such as "black man happy face", "Indian woman sad face", "Asian man angry face" etc. Collected images from these Google searches are checked for irrelevant images and tested for 80\% consensus in their annotations. By similar keyword searching on YouTube and other streaming platforms we collect relevant video clips from which we generate dataset images and labels. 

By aggregating these images from various sources we create our new dataset DeFi. We believe it is free of irrelevant images, more accurately annotated, more balanced in the number of samples in each class and represents people with a more diverse background. Each class in DeFi dataset has 7000+ images with 10\% images kept aside for the test set. Each image is $224\times224$ resolution. 

For all the images in DeFi, we create another dataset that imitates the face with the mask. This can be useful for other researchers who want to work with a dataset with masked people. To the best of our knowledge, any such dataset does not exist yet. The complete dataset will be publicly available at the provided web location~\cite{dd}.

We test \name \space in this new dataset. For both types of feature extractor of \name\space the accuracy is shown in the table~\ref{new_dataset}. 

\begin{table}[]
\caption{\name \space performance on proposed DeFi dataset}
\centering
\begin{tabular}{|l|l|}
\hline
 \textbf{Model}   &\textbf{Accuracy(\%)}  \\ 
 \hline
 CNN &  73.1\\
ResNet-50 &  78.3\\

\hline
\end{tabular}
\label{new_dataset}
\end{table}

\section{Conclusions and Future Work}
% In this paper, we presented \name, an improved system for emotion recognition from facial expressions. \name \space uses action units and visible properties of the face and state-of-the-art deep learning techniques to achieve higher accuracy. From the results of different experiments in the frequently used facial emotion recognition datasets, we showed \name \space has improved FER performance. Ablations studies showed the positive effects of background and place features on the accuracy. We also discussed some issues prevalent in many of the current datasets. We analysed the effect of mask-wearing during the Covid-19 pandemic on facial expression-based emotion recognition. Finally, we proposed a novel dataset for training with a section with masked subjects for the researcher community. In future, we plan to explore multi-modal emotion detection in situations where proper facial expression information is not available. We also want to explore more effective bias reduction techniques in facial emotion recognition. 

In conclusion, this paper presents \name, a novel system for emotion recognition from facial expressions that leverages action units, facial features, and state-of-the-art deep learning techniques. The system showed improved performance on benchmark datasets and demonstrated the positive effect of background and place features on recognition accuracy. Furthermore, the impact of mask-wearing during the Covid-19 pandemic on facial expression-based emotion recognition was analyzed and a novel dataset was proposed to address this novelty situation. In future work, we aim to investigate multi-modal techniques for novelty detection and mitigation in facial emotion recognition. These efforts promise to advance the state of the art in this field and have the potential to improve the robustness and effectiveness of facial emotion recognition systems.

\bibliography{main}

% \vskip -1\baselineskip plus -1fil

% \begin{IEEEbiography}[{\includegraphics[width=1in,height=1.25in,clip,keepaspectratio]{figs/Palash_us.jpg}}]{Mijanur Palash}
% received a BS degree in electrical and electronic engineering from Bangladesh University of Engineering and Technology and an MS in electrical and computer engineering from Southern Illinois University Carbondale, IL. Currently, he is working toward PhD in computer science at Purdue University, Indiana, USA. His research interests include multi-modal deep learning. 
% \end{IEEEbiography}

% \vskip -3\baselineskip plus -1fil

% \begin{IEEEbiography}[{\includegraphics[width=1in,height=1.25in,clip,keepaspectratio]{figs/bb.jpg}}]{Bharat Bhargava}
% is a professor in the Department of Computer Science at Purdue University. He is a fellow of the Institute of Electrical and Electronics Engineers. In 1999, he received the IEEE technical achievement award. He is the founder of the IEEE Symposium on Reliable and Distributed Systems, the IEEE conference on Digital Library, and the ACM Conference on Information and Knowledge Management. He received his PhD in Electrical engineering from Purdue University in 1974. His current research interest includes intelligent autonomous system, data analytics and machine learning. 
% \end{IEEEbiography}

\end{document}